# Sparse Topical Coding

**Jun Zhu, Eric P. Xing**
School of Computer Science
Carnegie Mellon University
Pittsburgh, PA 15213

## Abstract

We present *sparse topical coding* (STC), a non-probabilistic formulation of topic models for discovering latent representations of large collections of data. Unlike probabilistic topic models, STC relaxes the normalization constraint of admixture proportions and the constraint of defining a normalized likelihood function. Such relaxations make STC amenable to: 1) directly control the sparsity of inferred representations by using sparsity-inducing regularizers; 2) be seamlessly integrated with a convex error function (e.g., SVM hinge loss) for supervised learning; and 3) be efficiently learned with a simply structured coordinate descent algorithm. Our results demonstrate the advantages of STC and supervised MedSTC on identifying topical meanings of words and improving classification accuracy and time efficiency.

## 1 Introduction

Learning a representation that captures the latent semantics of a large collection of data is important in many scientific and engineering applications. Probabilistic topic models (PTM) such as latent Dirichlet allocation (LDA) [2] posits that each document is an admixture of latent topics where each topic is a unigram distribution over the terms in a vocabulary. The document-specific admixture proportion can be regarded as a representation of the document in the topic space, which can be used for classification or retrieval; and the inferred word-level topic assignment distributions can be used for word sense induction [3]. However, a limitation of such a PTM is that it lacks a mechanism to directly control the posterior sparsity [8] of the inferred representations. Sparsity of the representations in a semantic space is desirable in text modeling [21, 25] and human vision [20]. For example, very often it makes intuitive sense to assume that each document or word has a few salient topical meanings or senses [21, 25], rather than letting every topic make a non-zero contribution; this is important in practice for large scale text mining endeavors such as those undertaken in Google or Yahoo, where it is not uncommon to learn hundreds or thousands of topics for hundreds of millions of documents – without an explicit sparcification procedure, it would be extremely challenging, if not impossible, to nail down the semantic meanings of a document or word.

To achieve sparsity in a PTM is non-trivial. Existing attempts by using a sparse prior (e.g., Dirichlet [2]) or introducing auxiliary variables [25] could *indirectly* introduce a sparsity bias over the posterior representations. An arguably better way is to *directly* impose posterior regularization (e.g., posterior regularization using moment constraints [8] or entropic priors [21]). However, due to the smoothness of the regularizer (e.g., entropic regularizer), such methods often do not yield truly sparse posterior representations in practice.

A technical reason for the difficulty in achieving sparsity in PTMs is that the admixing proportions or topics are normalized distributions. Therefore, it is unhelpful to directly use a sparsity inducing $\ell_1$-regularizer as in lasso [24]. In contrast, the non-probabilistic sparse coding (SPC) [20] or non-negative matrix factorization (NMF) [16, 9] provides an elegant framework to achieve sparsity on the usually unnormalized code vector or dictionary by using the theoretically sound $\ell_1$-regularizer or other composite regularizers [13, 12, 1]. Due to the same reason of having to define a normalized likelihood function, another limitation of a PTM is that it usually has to deal with a hard-to-compute *log-sum-exp* function when considering discrete side information, such as label categories [26] and rich conditional features [30].

To address the above limitations, we present *sparse topical coding* (STC), a *non-probabilistic* formulation of topic models for learning hierarchical latent representations of input samples (e.g., text documents). In STC, each individual input feature (e.g., a word count) is *reconstructed* from a linear combination of a set of bases, where the coefficient vectors (or codes) are unnormalized, and the representation of an entire document is derived via an aggregation strategy (e.g., truncated averaging) from the codes of all its in-dividual features. When applied to text, we use the log-Poisson loss to model discrete word counts and learn the topical bases that are unigram distributions over the terms in a vocabulary. The relaxed non-probabilistic STC enjoys three nice properties which make it an appealing alternative formulation of topic models: 1) by imposing appropriate regularizers, STC can directly control the sparsity of the inferred representations; 2) the learning problem can be efficiently solved with a coordinate descent algorithm, which has closed-form solutions for updating code vectors; 3) STC can be seamlessly integrated with any convex loss function, which does not necessarily arise from a normalized probabilistic model, to incorporate supervised side-information for discovering *predictive* representations. Specifically, we describe a supervised MedSTC that integrates STC with a large-margin hinge-loss for considering categorical labels. Due to the non-probabilistic nature, MedSTC avoids dealing with an annoying normalization factor, which can make a supervised PTM [26] hard to do inference and learning. Our empirical studies show that: 1) STC can learn meaningful topical bases and identify sparse topical senses of words; and 2) both STC and MedSTC outperform several competing methods on document classification and are significantly more efficient (an order of magnitude speed up) on training and testing.

**Related work**: Although much work has been done on learning a structured dictionary [13, 1], existing SPC typically discovers flat representations, such as single-layer sparse codes of image patches or word terms [13, 1]. In order to achieve a representation of an entire document, post-processing such as average or max pooling [27] is needed. This two-step procedure can be rather sub-optimal because it lacks a channel to directly correlate individual component representations [11], or to leverage the possibly available supervision (e.g., document categories) to discover predictive representations [28] or learn a supervised dictionary [18]. NMF [16] uses one document-specific code vector to reconstruct *all* the word counts in the same document. This assumption is too limiting to capture the sparse topical meanings of each individual word. STC generalizes both SPC and NMF to discover hierarchical topical representations and allows different words in one document to exhibit different sparsity patterns via using different word codes (please see Appendix A.5 and Sec 2.4.3 for more details).

# 2 Sparse Topical Coding

Let $V = \{1, \cdots, N\}$ be a vocabulary with $N$ words. We represent a document as a vector $\mathbf{w} = (w_1, \cdots, w_{|I|})^\top$, where $I$ is the index set of words that appear and each $w_n$ ($n \in I$) represents the number of appearances of word $n$ in this document. Let $\boldsymbol{\beta} \in \mathbb{R}^{K \times N}$ be a dictionary with $K$ bases. We assume that each row $\boldsymbol{\beta}_k$ is a topic basis, i.e., a unigram distribution over $V$. Let $\mathcal{P}$ be a $(N\text{–}1)$-simplex, then $\boldsymbol{\beta}_k \in \mathcal{P}$. We will use $\boldsymbol{\beta}_{.n}$ to denote the $n$th column of $\boldsymbol{\beta}$. Then we present STC as a technique to project the input $\mathbf{w}$ into a semantic space that is spanned by a set of automatically learned bases $\boldsymbol{\beta}$ and achieve a high-level representation of the entire document jointly. Graphically, STC is a hierarchical latent variable model, as shown in Fig. 1, where $\boldsymbol{\theta}_d \in \mathbb{R}^K$ is the *document code* of document $d$ and $\mathbf{s}_{dn} \in \mathbb{R}^K$ is the *word code* of word $n$. Both document codes and word codes can be used for many tasks [2, 3].

In STC, we are particularly interested in learning sparse latent representations. We formulate STC as regularized loss minimization [24, 20, 13]. However, purely for the ease of understanding, we start with describing a probabilistic generative procedure.

## 2.1 A Probabilistic Generative Process

For simplicity, we assume that for each document the word codes $\mathbf{s}_n$ are conditionally independent given its document code $\boldsymbol{\theta}$ and the observed word counts are independent given their latent representations $\mathbf{s}$. We first sample a dictionary $\boldsymbol{\beta}$ from a uniform distribution[1] on $\mathcal{P}$. Then, each document can be described as arising from the following process

1. sample the document code $\boldsymbol{\theta}$ from a prior $p(\boldsymbol{\theta})$.
2. for each observed word $n \in I$
   (a) sample the word code $\mathbf{s}_n$ from a conditional distribution $p(\mathbf{s}_n|\boldsymbol{\theta})$
   (b) sample the observed word count $w_n$ from a distribution with the mean being $\mathbf{s}_n^\top \boldsymbol{\beta}_{.n}$.

The idea is that we treat $\mathbf{s}_n$ as a coefficient vector and use the linear combination $\mathbf{s}_n^\top \boldsymbol{\beta}_{.n}$ to reconstruct the observed word count $w_n$, under some loss measure as explained below; and the document code $\boldsymbol{\theta}$ is obtained via an aggregation of the individual codes of all its terms. The aggregation strategy depends on the choices of $p(\boldsymbol{\theta})$ and $p(\mathbf{s}|\boldsymbol{\theta})$, which also reflect our bias on the discovered representations. We will discuss them in the next section. For the last step of generating observed word counts, we adopt the broad class of exponential family distributions to make STC applicable to rich forms of data. Formally, we use the linear combination $\mathbf{s}_n^\top \boldsymbol{\beta}_{.n}$ as the *mean* parameter of an exponential family distribution that generates the observations $w_n$. In other words, we find an exponential family distribution $p(w_n|\mathbf{s}_n, \boldsymbol{\beta})$ that satisfies

$$\mathbb{E}_{p(w_n|\mathbf{s}_n,\boldsymbol{\beta})}[T(w_n)] = \mathbf{s}_n^\top \boldsymbol{\beta}_{.n}, \quad (1)$$

where $T(w_n)$ are the sufficient statistics[2] of $w_n$. We note that [17] uses the similar linear combination as the *natural* parameter of an exponential family distribution. We choose to use it as mean parameter because: 1) it is natural to constrain the feasible domains (e.g., non-negative for modeling word counts) of word codes for good interpretation, as explained below; and 2) in many cases, such as Poisson, Bernoulli and Gaussian, the distribution is commonly expressed with mean parameters. [4] uses a similar method as ours in defining exponential family distributions.

[1] Using a sophisticated prior is our future study.

[2] In general, $\mathbf{s}_n$ will be a matrix if $T$ is a vector.

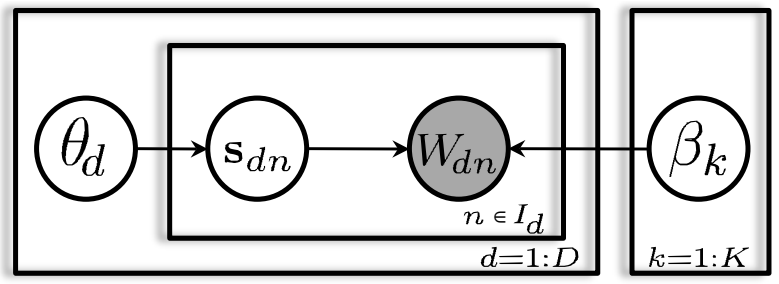

Figure 1: A two layer sparse topical coding model.

## 2.2 STC for Sparse MAP Estimation

Now, we formally define STC as finding the MAP estimate of the above probabilistic model, under a bias towards finding sparse latent representations.

The above generating procedure defines a joint distribution $p(\boldsymbol{\theta}, \mathbf{s}, \mathbf{w}|\boldsymbol{\beta}) = p(\boldsymbol{\theta})\prod_{n \in I} p(\mathbf{s}_n|\boldsymbol{\theta})p(w_n|\mathbf{s}_n, \boldsymbol{\beta})$. For discrete word counts, we use the Poisson distribution to generate the observations, i.e., $p(w_n|\mathbf{s}_n, \boldsymbol{\beta}) = Poiss(w_n; \mathbf{s}_n^\top \boldsymbol{\beta}_{.n})$, where $Poiss(x;\nu) = \frac{\nu^x e^{-\nu}}{x!}$. In order to achieve sparse codes $\boldsymbol{\theta}$ and $\mathbf{s}$, we choose the Laplace prior $p(\boldsymbol{\theta}) \propto \exp(-\lambda\|\boldsymbol{\theta}\|_1)$, and we define $p(\mathbf{s}_n|\boldsymbol{\theta})$ as a composite distribution $p(\mathbf{s}_n|\boldsymbol{\theta}) \propto \exp(-\gamma\|\mathbf{s}_n - \boldsymbol{\theta}\|_2^2 - \rho\|\mathbf{s}_n\|_1)$, which is supergaussian [10]. The $\ell_1$-norm will bias towards finding sparse codes. The hyper-parameters $(\lambda, \gamma, \rho)$ are non-negative and they can be selected via cross-validation or integrated out by introducing hyper-priors [5, 7].

Let $\Theta = \{\boldsymbol{\theta}_d, \mathbf{s}_d\}_{d=1}^D$ denote the codes for a collection of documents $\{\mathbf{w}_d\}_{d=1}^D$. STC solves the problem

$$\min_{\Theta, \boldsymbol{\beta}} \sum_{d,n \in I_d} \ell(\mathbf{s}_{dn}, \boldsymbol{\beta}) + \lambda \sum_d \|\boldsymbol{\theta}_d\|_1 + \sum_{d,n \in I_d} (\gamma\|\mathbf{s}_{dn} - \boldsymbol{\theta}_d\|_2^2 + \rho\|\mathbf{s}_{dn}\|_1)$$
$$\text{s.t.}: \boldsymbol{\theta}_d \geq 0,\ \forall d;\ \ \mathbf{s}_{dn} \geq 0,\ \forall d, n \in I_d;\ \ \boldsymbol{\beta}_k \in \mathcal{P},\ \forall k, \qquad (2)$$

where the objective function is the negative logarithm of the posterior $p(\Theta, \boldsymbol{\beta}|\{\mathbf{w}_d\})$ with a constant omitted and $\ell$ is a loss function. For text, we have $\ell(\mathbf{s}_n, \boldsymbol{\beta}) = -\log Poiss(w_n; \mathbf{s}_n^\top \boldsymbol{\beta}_{.n})$. Minimizing the log-Poisson loss is actually equivalent to minimizing an unnormalized KL-divergence between observed word counts $w_n$ and their reconstructions $\mathbf{s}_n^\top \boldsymbol{\beta}_{.n}$ [23]. Since word counts are non-negative, a negative $\boldsymbol{\theta}$ or $\mathbf{s}$ will lose interpretability. Therefore, we constrain $\boldsymbol{\theta}$ and $\mathbf{s}$ to be non-negative. A non-negative code can be interpreted as representing the relative importance of topics. Moreover, as shown in [16], imposing non-negativity constraints could potentially result in sparser and more interpretable patterns.

## 2.3 Optimization with Coordinate Descent

Let $f(\Theta, \boldsymbol{\beta})$ denote the objective of problem (2). When using a convex loss function $\ell$ (e.g., log-loss under the exponential family of distributions), $f(\Theta, \boldsymbol{\beta})$ is generally bi-convex, that is, convex over either $\Theta$ or $\boldsymbol{\beta}$ when the other one is fixed. Moreover, the feasible set is a convex set. Therefore, a natural algorithm to solve this bi-convex problem is coordinate descent, as typically used in sparse coding methods [17, 1]. Specifically, the procedure alternatively performs:

**Hierarchical sparse coding**: this step involves finding the codes $\Theta$ when $\boldsymbol{\beta}$ is fixed. Due to the conditional independency, we can perform this step for each document separately by solving the convex problem

$$\min_{\boldsymbol{\theta}, \mathbf{s}} \sum_{n \in I} \ell(\mathbf{s}_n, \boldsymbol{\beta}) + \lambda\|\boldsymbol{\theta}\|_1 + \sum_{n \in I} (\gamma\|\mathbf{s}_n - \boldsymbol{\theta}\|_2^2 + \rho\|\mathbf{s}_n\|_1)$$
$$\text{s.t.}: \boldsymbol{\theta} \geq 0;\ \ \mathbf{s}_n \geq 0,\ \forall n \in I.$$

While previous work used local quadratic approximation to achieve a lasso type problem [17] or specialized Poisson likelihood estimation [23], we solve this problem with coordinate descent, which has a closed-form solution for each component of $\mathbf{s}$ and $\boldsymbol{\theta}$. Moreover, our method offers an algorithmic comparison with LDA, as discussed later. Specifically, we alternatively solve:

*Optimize over* $\mathbf{s}$: when $\boldsymbol{\theta}$ is fixed, $\mathbf{s}_n$ are not coupled. For each $\mathbf{s}_n$, we solve the problem

$$\min_{\mathbf{s}_n} \ell(\mathbf{s}_n, \boldsymbol{\beta}) + \gamma\|\mathbf{s}_n - \boldsymbol{\theta}\|_2^2 + \rho\sum_k s_{nk},\ \text{s.t.}: \mathbf{s}_n \geq 0,$$

where we have explicitly written the $\ell_1$-norm of $\mathbf{s}_n$ under the non-negativity constraint. Let $g(\mathbf{s}_n)$ be the objective. Then, we solve for each $s_{nk}$ alternatively. By Proposition 1, as to be presented, the solution is $s_{nk} = \max(0, \nu_k)$, where $\nu_k = \arg\min_{s_{nk}} g(\mathbf{s}_n)$ with $s_{nj},\ j \neq k$ fixed at current solutions. By setting the gradient $\nabla_{s_{nk}} g = (1 - \frac{w_n}{\mathbf{s}_n^\top \boldsymbol{\beta}_{.n}})\beta_{kn} + 2\gamma(s_{nk} - \theta_k) + \rho$ equal to zero, we have that $\nu_k$ is the solution of the equation

$$2\gamma\beta_{kn}\nu_k^2 + (2\gamma\mu + \beta_{kn}\tau)\nu_k + \mu\tau - w_n\beta_{kn} = 0,$$

where $\mu = \sum_{j \neq k} s_{nj}\beta_{jn}$ and $\tau = \beta_{kn} + \rho - 2\gamma\theta_k$. If $\beta_{kn} = 0$, we have $\nu_k = \theta_k - \frac{\rho}{2\gamma}$; otherwise, we need to solve a quadratic equation, which always has real solutions because the discriminant $\nabla \triangleq (2\gamma\mu + \beta_{kn}\tau)^2 - 4(2\gamma\beta_{kn})(\mu\tau - w_n\beta_{kn}) = (2\gamma\mu - \beta_{kn}\tau)^2 + 8\gamma w_n \beta_{kn}^2$ is guaranteed to be positive. $\nu_k$ is the larger one of the two possible solutions.

*Optimize over* $\boldsymbol{\theta}$: when $\mathbf{s}$ is fixed, this step involves solving the convex problem

$$\min_{\boldsymbol{\theta}} \lambda\|\boldsymbol{\theta}\|_1 + \gamma\sum_{n \in I}\|\mathbf{s}_n - \boldsymbol{\theta}\|_2^2,\ \text{s.t.}: \boldsymbol{\theta} \geq 0.$$

Since different dimensions of $\boldsymbol{\theta}$ are not coupled, we can solve for each $\theta_k$ separately. By Proposition 1, we have

$$\forall k,\ \theta_k = \max(0, \bar{s}_k - \frac{\lambda}{2\gamma|I|}), \qquad (3)$$

where $\bar{s}_k = \frac{1}{|I|}\sum_{n \in I} s_{nk}$. Therefore, using an $\ell_1$-regularizer gives us a *truncated averaging*[3] strategy for aggregating individual word codes to obtain $\boldsymbol{\theta}$.

**Dictionary learning**: after we have inferred the latent representations $(\boldsymbol{\theta}, \mathbf{s})$ of all the documents, we update the dictionary $\boldsymbol{\beta}$ by minimizing the log-Poisson loss, which is convex and can be efficiently solved with a high-performance method, such as projected gradient descent, where the projection to the simplex $\mathcal{P}$ can be performed with a linear algorithm [6].

### 2.4 Discussions

Now, we investigate some properties of STC.

#### 2.4.1 Generality

According to the following proposition, the coordinate descent algorithm is generally applicable for any convex loss function $\ell$, e.g., log-loss of exponential family distributions. For different loss functions, the difference lies in solving a univariate minimization problem for each $s_{nk}$, which can have a closed-form solution in some cases such as Poisson and Gaussian. In general, this univariate problem can be efficiently solved with a numerical method if no closed-form solution exists.

**Proposition 1** *Let $h(x)$ be a strictly convex function. The optimum solution $x^\star$ of the constrained problem $P_0: \min_{x \geq 0} h(x)$ is $x^\star = \max(0, x_0)$, where $x_0$ is the solution of the unconstrained problem $P_1: \min_x h(x)$.*

*Proof:* See Appendix A.1 for details. □

#### 2.4.2 Connections to Probabilistic LDA

STC is a non-probabilistic formulation of topic models. Now, we provide a systematical comparison with the representative probabilistic topic model – LDA [2].

First, LDA doesn't have an explicit definition of word code. In LDA, a document is represented as a *sequence* $\mathbf{w} = (\vec{w}_1, \cdots, \vec{w}_M)$, where $M$ is document length and $\vec{w}_m$ is an $N$-dim indicator vector (i.e., $w_{mn} = 1$ if word $n$ appears in position $m$, otherwise 0). LDA associates each position $m$ with a topic assignment variable $Z_m$ and assumes that the topics of all the words in a document are sampled from the same document-specific mixing proportion (denoted by $\boldsymbol{\theta}$ too), which has a Dirichlet prior $p(\boldsymbol{\theta}|\alpha)$. For comparison, an equivalence to word code can be defined as the *empirical* word-topic assignment distribution $\tilde{p}(z(n) = k) \propto \sum_m w_{mn} p(z_{mk} = 1|\mathbf{w})$, where $z(n)$ is the topic of word $n$. The distribution $\tilde{p}(z(n))$ can be regarded as a representation of word $n$ in the topic space, and it can be inferred using sampling [3] or variational [2] methods.

Second, LDA lacks an explicit sparcification procedure on the inferred representations as discussed in Sec 1. Although we can adjust $\alpha$ to make $\boldsymbol{\theta}$ concentrate much of its mass on a small number of topics *a priori*, it can only indirectly influence the sparsity of inferred posterior representations [8]. In practice, using a Dirichlet prior is not effective in controlling the posterior sparsity of LDA models. Fig. 2(L) shows the *sparsity ratio* of word code (i.e., number of zeros in the code divided by topic number $K$) and classification accuracy (see Sec 4.2) with different pre-specified Dirichlet parameter $\alpha$ of LDA using variational inference[4]. We can see that a small $\alpha$ (i.e., a weak Dirichlet smoothing [2]) can yield sparse representations because of data scarcity, but this sparsity is not good for classification. Using a large $\alpha$ (i.e., a strong Dirichlet smoothing) can increase the accuracy, but it dramatically reduces the sparsity ratio. Also, there is a sharp change point around $\alpha = 10^{-3}$.

Finally, the learning algorithm of STC has the similar structure as the variational EM algorithm of LDA [2], as outlined in Appendix A.2. The difference lies in: 1) STC is doing deterministic coordinate descent while LDA performs probabilistic inference under normalization constraints; and 2) STC uses projected gradient descent to solve for the topics $\boldsymbol{\beta}$ while LDA performs this step in a closed-form. Empirically, as we shall see in Sec 4.3, STC is much more efficient than LDA on inferring latent representations and the overall training time of STC is also much smaller than that of LDA because the dictionary learning step is much faster than the hierarchical sparse coding step, especially when the number of samples is large.

#### 2.4.3 Comparison with SPC and NMF

As we have stated, STC is an extension of SPC and NMF for discovering hierarchical representations with the bases being distributional topics. Another difference is that STC only encodes the words with nonzero counts, while the SPC [17] and standard NMF encode all the words in a vocabulary. This difference could make STC more efficient and scalable to a large vocabulary. See Appendix A.5 for more comparison.

## 3 Supervised Sparse Topical Coding

We have described the *unsupervised* STC for learning dictionary and inferring sparse representations of un-labeled samples. But with the increasing availability of free on-line information such as image tags, user ratings, etc., it is desirable to develop new models and training schemes that can make effective use of such "free" supervised side information to achieve better results, such as more discriminative latent representations of text contents, and more accurate classifiers.

[3]If we use a normal prior $p(\boldsymbol{\theta}) \propto \exp(-\lambda \|\boldsymbol{\theta}\|_2^2)$, we will have $\boldsymbol{\theta}_k = \frac{\gamma}{\lambda/|I| + \gamma} \bar{s}_k$. If $\lambda \ll \gamma$, $\boldsymbol{\theta}$ will be close to the *averaging* aggregation of its individual word codes. Another choice is to set $\lambda = \gamma$, and we have $\theta_k = \frac{|I|}{1+|I|} \bar{s}_k$, which is again close to the average if $|I|$ is large.

[4]In theory, variational methods don't produce zero code elements because of the exponential update rule. But in practice, it is safe to truncate very small values to be zero. Similarly, sampling methods don't have a direct control on the posterior sparsity either.

Now, we present a *supervised* STC to learn predictive representations and a supervised dictionary [18] by exploring the available side-information. We consider the classification problem, where the response variable $Y$ takes a value from a finite set of categorical labels. As we mentioned, the non-probabilistic STC can be naturally integrated with any convex loss function, which may or may not arise from a probabilistic model. Here, we adopt the large margin principle to define a classifier, which can avoid dealing with a normalization factor as involved in probabilistic models [26] and has been successfully explored in MedLDA [28].

Specifically, we use document code $\boldsymbol{\theta}$ as input features for a large margin classifier, and define the linear discriminant function $F(y, \boldsymbol{\theta}) = \boldsymbol{\eta}_y^\top \boldsymbol{\theta}$, where $\boldsymbol{\eta}_y \in \mathbb{R}^K$. Let $\boldsymbol{\eta}$ denote the set of $\boldsymbol{\eta}_y$, and let $\Delta\ell(y, y')$ be a cost function that measures how different a prediction $y'$ is from the true label $y$. Then, given a training set $\mathcal{D} = \{(\mathbf{w}_d, y_d)\}_{d=1}^D$, the multi-class hinge loss is $\mathcal{R}_h(\{\boldsymbol{\theta}_d\}, \boldsymbol{\eta}) = \frac{1}{D}\sum_d \max_y[\Delta\ell(y_d, y) + F(y, \boldsymbol{\theta}_d) - F(y_d, \boldsymbol{\theta}_d)]$. We define the max-margin supervised STC (MedSTC) as jointly learning a large-margin classifier $\boldsymbol{\eta}$, learning a dictionary $\boldsymbol{\beta}$, and discovering latent representations $\Theta$. The joint optimization problem is

$$\min_{\Theta, \boldsymbol{\beta}, \boldsymbol{\eta}} \; f(\Theta, \boldsymbol{\beta}) + C\mathcal{R}_h(\{\boldsymbol{\theta}_d\}, \boldsymbol{\eta}) + \frac{1}{2}\|\boldsymbol{\eta}\|_2^2 \qquad (4)$$

$$\text{s.t.}: \; \boldsymbol{\theta}_d \geq 0, \; \forall d; \; \mathbf{s}_{dn} \geq 0, \; \forall d, n \in I_d; \; \boldsymbol{\beta}_k \in \mathcal{P}, \; \forall k,$$

where $C$ is a positive constant. We can see that the document code $\boldsymbol{\theta}$ plays a role of bridging the internal latent representations to the external supervision.

For problem (4), we can use the similar coordinate descent method, with slight changes on solving for $\boldsymbol{\theta}$ and an additional step for learning $\boldsymbol{\eta}$. For $\boldsymbol{\eta}$, the problem is to learn a multi-class SVM, which can be done with an existing solver [14]; for $\boldsymbol{\theta}$, we can again achieve its optimum as in Eq. (3) with $\bar{s}_k$ being a *shifted mean*. For document $d$, $\bar{s}_k = \frac{1}{|I_d|}\sum_{n \in I_d} s_{dnk} + \frac{C}{2D|I_d|\lambda}(\eta_{y_d k} - \eta_{\hat{y}_d k})$, where $\hat{y}_d = \arg\max_y(\Delta\ell(y_d, y) + F(y, \boldsymbol{\theta}_d))$ is the loss augmented prediction.

# 4 Experiments

Now, we provide qualitative as well as quantitative evaluation of STC and MedSTC on the 20 Newsgroups dataset, which contains 18775 postings in 20 categories. The vocabulary contains 61188 terms, and we remove a standard list of 524 stop words as in [15, 28].

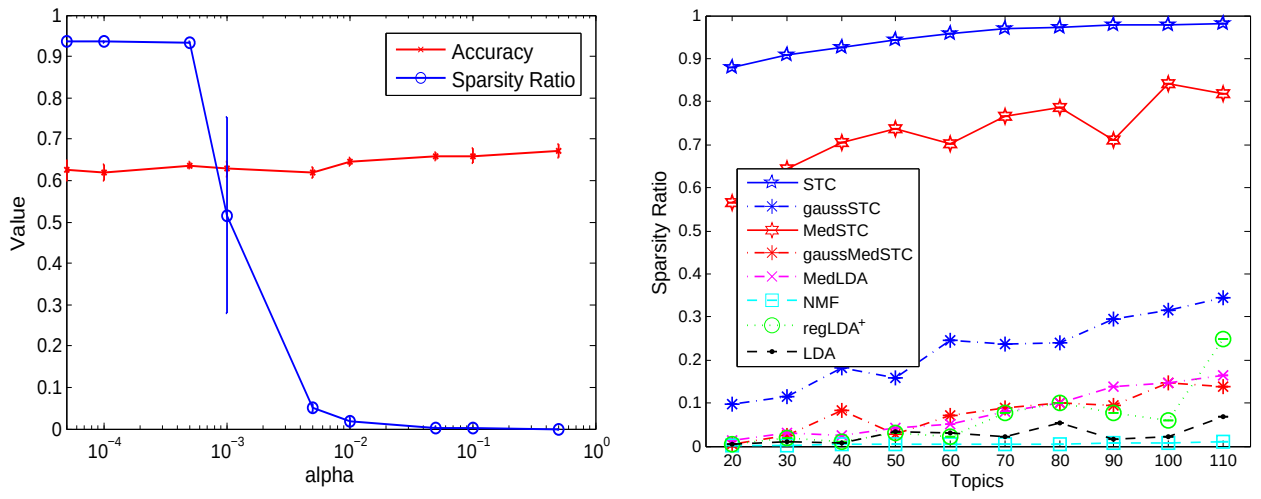

Figure 2: (L) sparsity ratio of word codes and classification accuracy of 70-topic LDA with different pre-specified $\alpha$ of a symmetric Dirichlet prior; and (R) sparsity ratio of word codes discovered by different models. For LDA models, we estimate the optimal Dirichlet parameter $\alpha$.

We observe that using $\ell_2$-norm on $\theta$ leads to sparser word codes, but denser document codes. We report the results of STC and MedSTC using $\ell_2$-norm on $\theta$ and leave other results to Appendix. To see the effects of $\ell_1$-norm, we also report the performance of *gaussSTC* (i.e., an STC that uses $\ell_2$-norm on both $\theta$ and $\mathbf{s}$). The supervised guassSTC is denoted by *gaussMedSTC*.

## 4.1 Characteristics of Code Representation

**Word code**: Fig. 3 shows the average word codes of some representative words[5] in 6 example categories. Here, we compare STC with LDA [2] using variational inference. The Dirichlet parameter $\alpha$ in LDA is automatically estimated using the Newton-Raphson method [2]. For each word $n$, we compute the average code weights $\bar{s}_{nk} = \frac{1}{|\mathcal{D}_n|}\sum_{d \in \mathcal{D}_n} s_{dnk}$ over *all* the documents (indexed by $\mathcal{D}_n$) that word $n$ appears in. For LDA, the average code of word $n$ is $\bar{p}(z(n)) = \frac{1}{|\mathcal{D}_n|}\sum_{d \in \mathcal{D}_n} \tilde{p}(z_d(n))$ (See Sec 2.4.2 for definition of $\tilde{p}(z(n))$). For each category, we show the topics learned by STC that have non-zero weights on at least one representative word. We can see that the codes discovered by STC are much sparser than those discovered by LDA. For STC, on average, each word has a small number of non-zero code elements, all of which are significantly larger than zero. In contrast, the word codes discovered by LDA tend to have many small non-zeros, as also characterized by the sparsity ratio in Fig. 2. LDA does have sparse codes for some words (e.g., *jpeg*), but the sparsity is mainly due to data scarcity, as we have analyzed in Sec 2.4.2.

By closely examining the learned topics, we can see that in STC each non-zero element in the code of a word roughly represents one of its topical meanings. For instance, the word *speed* has non-zero code values on topics *T14*, *T17*, *T18* and *T19*, which are roughly about hardware speed, circuit speed (e.g., scsi bus speed), drive speed, and the cost/price on system speed, respectively. The weights on these topics reflect their relative significance in the corpus. We can also see that some words (e.g., *religion* and *medical*) have only one or two topical meanings, while some other words (e.g., *speed* and *system*) tend to have a broad spectrum of topical meanings.

[5] We choose 3 words that most frequently appear in the documents in each category (independent of models). To be diverse, overlapping words are avoided.

| Class | STC | LDA | STC Topics |
|---|---|---|---|
| alt. atheism | 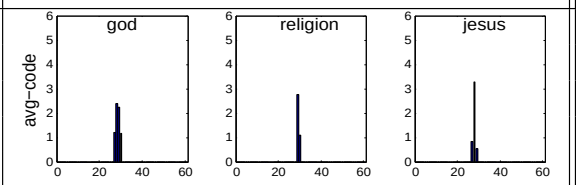 | 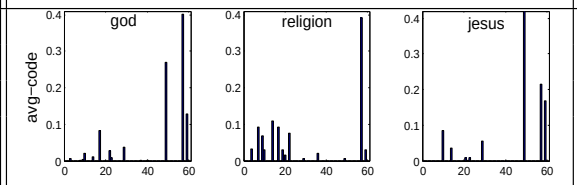 | T27: word, paul, argument, god, law, words, man, books, follow, language, context<br>T28: god, jesus, love, man, christ, john, son, bible, lord, christian, hell, sin<br>T29: god, church, truth, religion, christian, bible, rutgers, christianity, die, belief, faith<br>T30: life, science, evidence, exist, god, brian, created, existence, theory, claim, atheism |
| comp. graphics | 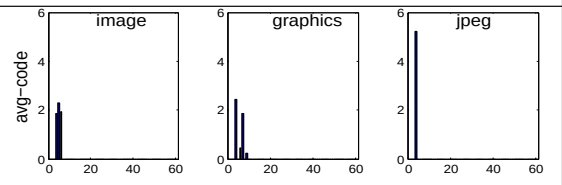 | 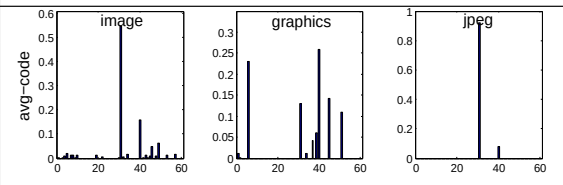 | T4: format, graphics, file, gif, ray, pub, jpeg, files, images, image, convert<br>T5: sun, color, motif, mit, window, image, server, display, dec, export, colors<br>T6: uk, window, create, type, manager, event, ac, application, server, image, program<br>T7: windows, dos, run, running, version, os, file, mode, files, software, mouse, pc<br>T9: mail, email, list, info, send, ftp, software, internet, network, contact, request |
| rec. autos | 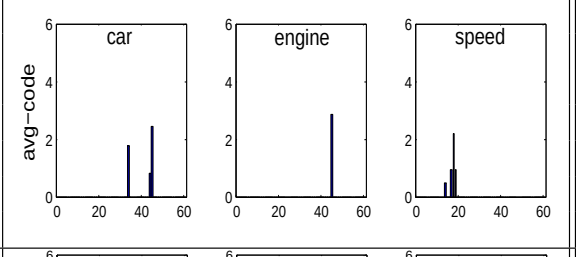 | 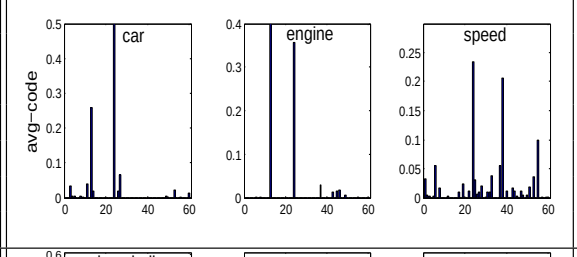 | T33: high, michael, light, friend, total, big, mine, month, lower, basically, cheap, costs<br>T44: water, hear, back, head, disclaimer, aren, room, told, don, door, chris, turn<br>T45: car, dod, bike, road, cars, engine, att, ride, miles, cb, rec, writes<br>T14: mac, apple, ibm, bit, video, port, card, mb, pin, hardware, se, internal<br>T17: scsi, ms, stephanopoulos, netcom, trust, bus, performance, clock, mb, faster, de<br>T18: drive, hard, card, drives, drivers, speed, system, scsi, board, mb, driver, disk<br>T19: local, cost, support, computer, system, buy, service, systems, cd, access, price |
| sport. baseball | 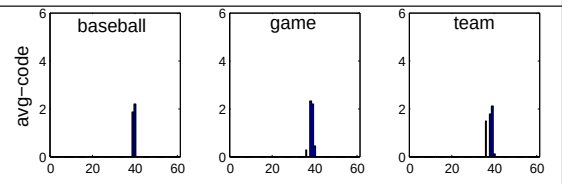 | 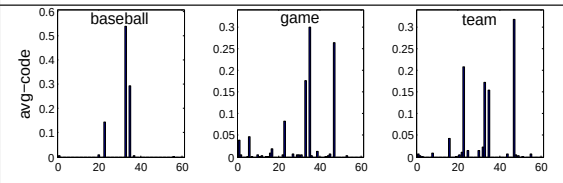 | T36: final, ca, peter, hockey, team, fi, bob, nhl, european, roger, draft, stars<br>T38: game, st, team, play, gm, cup, hockey, pick, division, round, chicago, pittsburgh<br>T39: year, games, win, game, team, san, red, series, record, season, york, joe<br>T40: year, runs, average, dave, field, hit, baseball, ball, jack, players, park, writes |
| sci. med | 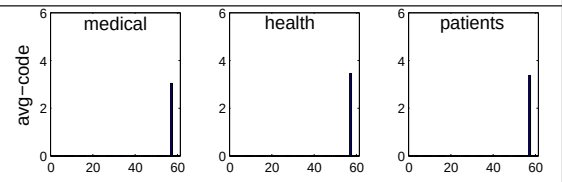 | 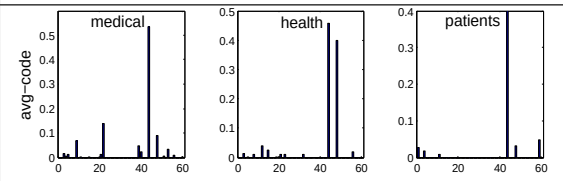 | T57: health, medical, study, pitt, age, dc, blood, research, disease, reported, doctor |
| comp.os. ms-windows. misc | 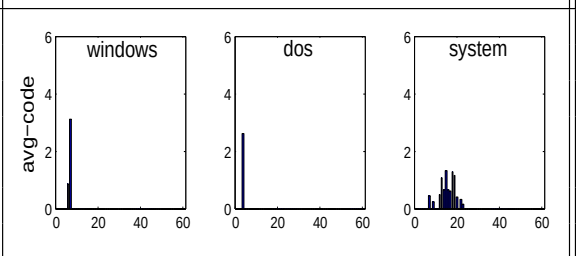 | 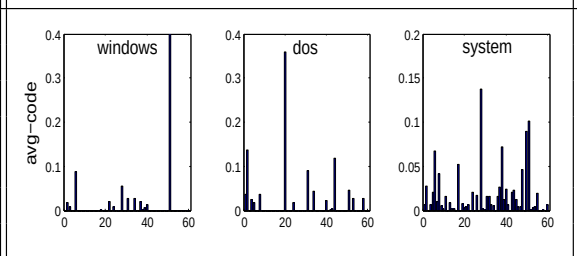 | T4: format, graphics, file, gif, ray, pub, jpeg, files, images, image, convert, ch<br>T6: uk, window, create, type, manager, event, ac, application, server, image, program<br>T7: windows, dos, run, running, version, os, file, mode, files, software, mouse<br>T13: bit, unit, machine, command, computer, system, standard, serial, number, data<br>T15: chip, phone, voice, system, data, standard, device, clipper, bit, writes, missing<br>T18: drive, hard, card, drives, drivers, speed, system, scsi, board, mb, driver<br>T19: local, cost, support, computer, system, buy, service, systems, cd, access, price |

Figure 3: Average word code of representative words for different categories discovered by STC and LDA models.

Fig. 2(R) shows the average *sparsity ratio* of different models on testing documents[6]. We can see that the word codes discovered by LDA (with estimated $\alpha$) and NMF[7] are very dense. In contrast, STC achieves much sparser codes by using the sparsity-inducing $\ell_1$-regularizer on word codes. For the similar reason, MedSTC is much sparser than MedLDA [28] and supervised LDA (sLDA) [26] whose sparsity is comparable to that of NMF and omitted for clarity. The reason why MedSTC is denser than STC is that response variables introduce additional correlations between different topics during inference, which lead to a spread of non-zero values. By comparing STC and gaussSTC, we can see that using the $\ell_1$-norm can significantly improve the sparsity ratio (and classification performance as we shall see). This suggests that using non-negativity constraints only is insufficient to achieve sparse representations. The similar observation applies to MedSTC and guassMedSTC.

We also compare with regularized LDA (regLDA) that uses an entropic regularizer [21] on the word-topic assignment distribution $p(z_{mk} = 1|\mathbf{w})$ during variational inference (See Appendix A.4). Fig. 2(R) shows the *regLDA*$^+$ that achieves the best classification performance. We can see that an entropic regularizer can bias LDA toward finding a sparser representation. But, it is not so effective as STC due to its smoothness. Moreover, as shown in Fig. 5 (*regLDA*$^-$), if we use a strong entropic regularizer to achieve the similar sparsity ratio in word code as STC, the classification performance will decrease dramatically.

| Class | STC | MedSTC |
|---|---|---|
| alt.atheism | 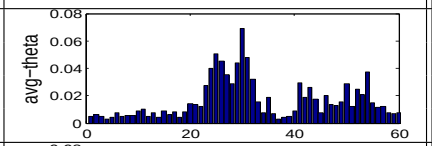 | 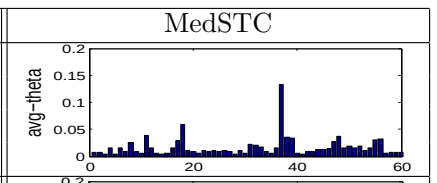 |
| misc.forsale | 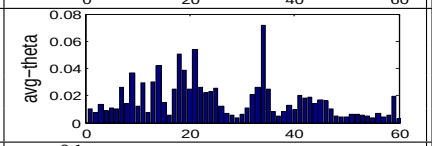 | 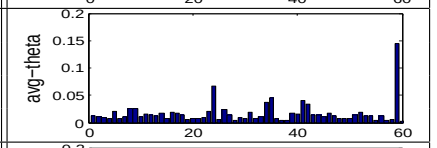 |
| rec.autos | 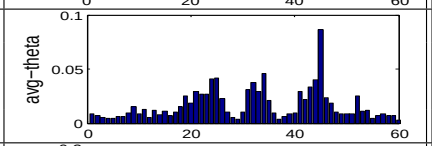 | 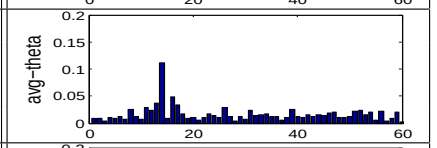 |
| sport.baseball | 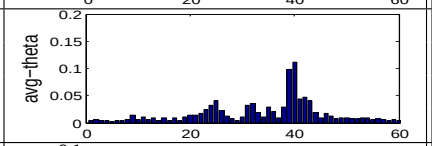 | 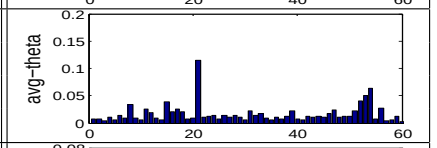 |
| sci.med | 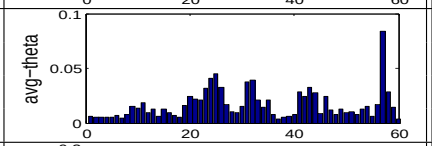 | 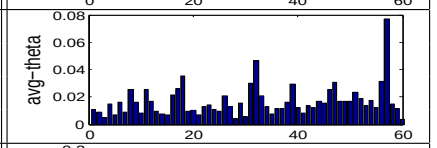 |
| religion.christian | 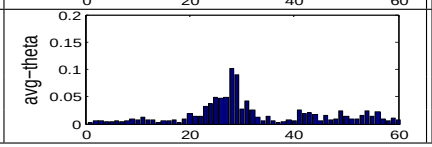 | 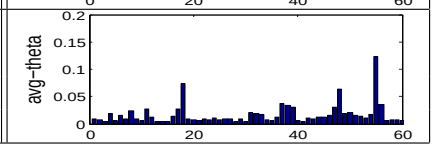 |

Figure 4: Average document code $\theta$ for example categories discovered by STC and MedSTC.

**Document code:** Fig. 4 shows the average document code $\theta$ for 6 example categories discovered by STC and MedSTC. For each category, we average $\theta$ over all the documents in that category and then normalize it. We can see that the average $\theta$ for different categories are quite different, which indicates that document codes have a good discriminative power. Moreover, by using supervised information, MedSTC can discover more discriminative representations. To save space, we omit the results of LDA, which is worse than STC as shown by classification accuracy. Note that the average $\theta$ is not sparse even though each document code $\theta_d$ can be sparse, especially when using $\ell_1$-norm on $\theta$).

[6] We use the same train/test split as in [15, 28].

[7] For NMF, word codes are the same as document codes. For improving efficiency, we ignore non-appearing words when implementing NMF. See Appendix A.5 for details.

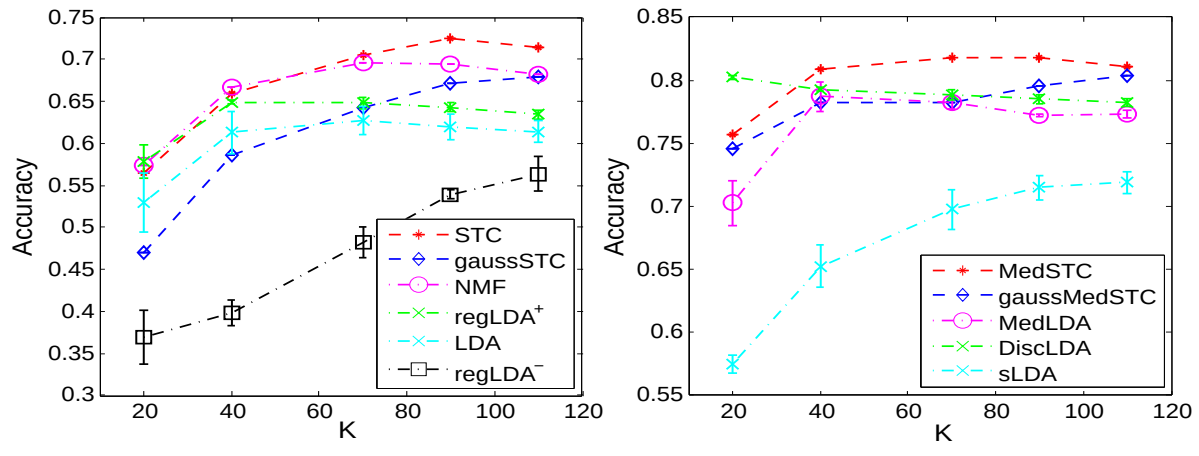

Figure 5: Classification accuracy of different models.

### 4.2 Prediction Accuracy

Now, we report the classification accuracy on the 20 Newsgroup data with all the 20 categories. We compare STC (and gaussSTC) with LDA, NMF and regLDA; and we compare MedSTC (and gaussMedSTC) with the max-margin based MedLDA and likelihood based sLDA and discriminative LDA (DiscLDA) [15]. For regLDA, we consider two versions – *regLDA*$^+$ (achieving best classification performance) and *regLDA*$^-$ (achieving similar sparsity in word code as STC). For unsupervised models, we use all the data to learn their parameters, including $\alpha$ in LDA models, and then use the training documents with their topical representations as features to build multi-class SVM classifiers. We use the same solver as in [28] with a cost function $\Delta\ell(y_d, y) \triangleq \ell\mathbb{I}(y \neq y_d)$ to solve the sub-step of learning $\boldsymbol{\eta}$ in MedSTC, and learn the SVM classifiers for STC, gaussSTC, NMF and LDA. For regularization constants, we set $\gamma = \lambda$, and perform cross-validation to select $C$, $\lambda$ and $\rho$ [8].

Fig. 5 shows the results with 5 randomly initialized runs. For STC models, we initialize $\boldsymbol{\beta}$ and $(\boldsymbol{\theta}, \mathbf{s})$ to be uniform. We can see that STC performs better than LDA and NMF, especially when $K$ is large. One possible reason for this improvement is that STC is flexible in discovering sparse code representations for each word. In contrast, the additional normalized constraint imposed on the word code (i.e., $\tilde{p}(z(n))$) may limit the flexibility of LDA to capture the intrinsic sparsity of topic meanings for an individual word, as shown in Fig. 2. NMF is limited too by using one code vector to reconstruct all the words in a document, as discussed in Sec 1. For supervised models, MedSTC performs better than MedLDA because of its similar sparsity property. The importance of learning sparse representations can also be seen from the inferior performance of gaussSTC (gaussMedSTC) compared to STC (MedSTC). Moreover, the max-margin model (e.g., MedSTC) generally outperform sLDA which is learned by using likelihood estimation. Finally, using an entropic regularizer in LDA (i.e., regLDA$^+$) can improve the accuracy a bit, but it is not so effective as STC, as indicated by regLDA$^-$ which achieves similar sparsity as STC but decreases accuracy dramatically.

[8]We set $\ell$ at 16 for MedLDA and 3600 for MedSTC. A reason for the difference is that MedLDA discovers normalized representations, while MedSTC generally discovers representations that are of a larger scale of magnitude.

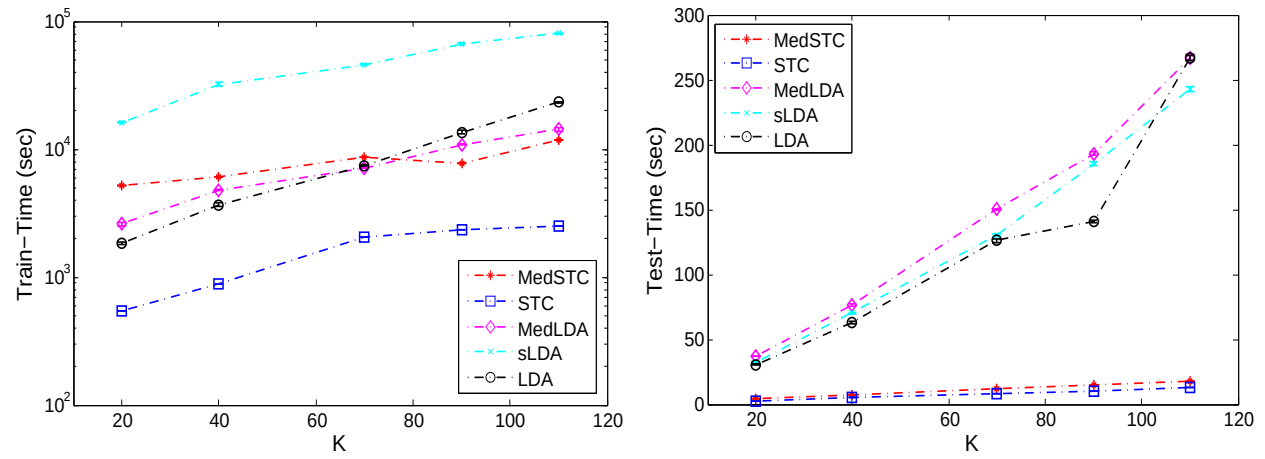

Figure 6: (L) training and (R) testing time.

### 4.3 Time Efficiency

All the models are implemented using the same data structures in C++ and run on a standard desktop with a 2.66GHz processor. To save space, we omit the result of gaussSTC (gaussMedSTC), which is comparable to that of STC (MedSTC). See Appendix A.5 for NMF.

Fig. 6 shows the training and test time of different topic models with 5 randomly initialized runs. For LDA and STC, training includes building SVM classifiers. Clearly, STC is much more efficient than LDA in training. The main reason is that the coordinate descent algorithm of STC is much more efficient than the probabilistic inference of LDA, as shown in Fig. 6(R). Moreover, the sparsity of word codes in STC can be utilized to further improve the efficiency (e.g., zeros aren't needed to be stored or multiplied). For the supervised MedSTC and MedLDA, which rely on a solver to learn a SVM, the training is mainly dependent on learning SVM as the inference is generally fast. Thus they have comparable training time, also comparable to LDA which usually needs more iterations to converge. Here, we use the *1-slack* formulation of multi-class SVM, which is faster than an equivalent *n-slack* formulation [14]. Among all the models, sLDA is the slowest one because it defines a normalized probabilistic model for the discrete variable $Y$, whose normalization factor (i.e., a *sum-exp* function) strongly couples the topic assignments of different words in the same document. Thus the posterior inference in training is much slower than that of LDA and MedLDA which uses LDA as the underlying topic model.

For testing, we can see that STC is much faster than probabilistic LDA. As compared in Sec 2.4.2, the main reason for this efficiency improvement is that the coordinate descent method is much more efficient than the variational inference procedure in LDA, which involves many calls to digamma functions [2] and needs an additional normalization step in order to get word-topic assignment distributions. For supervised models, MedSTC is much faster than MedLDA and sLDA, both of which use the same variational inference procedure as in LDA. The slight difference between MedLDA, sLDA and LDA is because they have different topics and the inference converges differently. DiscLDA is roughly about 20 times slower than MedLDA or sLDA in testing because it needs to infer the latent representations for each possible category.

## 5 Conclusions and Future Work

We have presented sparse topical coding (STC), an alternative non-probabilistic formulation of topic models for discovering latent representations of large collections of data. STC relaxes the normalization constraints made in probabilistic topic models. Such a relaxation makes STC enjoy nice properties, such as direct control on the sparsity of discovered representations, efficient learning algorithm, and seamless integration with a convex loss function for learning predictive latent representations. STC offers a systematical connection between sparse coding and probabilistic topic modeling. Our empirical studies demonstrate the advantages of STC and supervised MedSTC on identifying sparse topical meanings of words, and improving time efficiency and classification accuracy.

Due to the relaxation from defining normalized distributions, STC can efficiently incorporate rich features without dealing with annoying normalization factors, which can make the inference hard in probabilistic models [30]. We have extended STC to consider rich features and preliminary results are presented in [29]. For future work, we are interested in developing parallel STC for large-scale applications [19, 22], and we want to do a systematical study on automatically estimating the hyper-parameters [7].

Finally, the appendix and our code are available at http://www.cs.cmu.edu/∼junzhu/stc.htm.

**Acknowledgements**

We thank the reviewers for their valuable comments. This work is supported by AFOSR FA95501010247, ONR N000140910758, NSF Career DBI-0546594, and an Alfred P. Sloan Research Fellowship to EPX.